\documentclass[journal]{IEEEtran}
\usepackage{graphicx}
\usepackage{amsmath}
\usepackage{epstopdf}
\usepackage{amssymb}
\usepackage[ruled,vlined,commentsnumbered]{algorithm2e}
\usepackage{algpseudocode}
\usepackage{array}
\usepackage[]{algorithm2e}
\usepackage{booktabs}
\usepackage{multirow}
\usepackage{amsfonts}
\usepackage{amsbsy}
\usepackage{wasysym}
\usepackage{times}
\usepackage{booktabs, threeparttable, stackengine}
\usepackage{array}
\usepackage{color}
\usepackage{tabularx}
\usepackage[utf8]{inputenc}
\usepackage[T1]{fontenc}
\usepackage{lmodern}
\usepackage[figurename=Fig.,labelfont=bf,labelsep=period]{caption}
\usepackage{amsmath}
\usepackage{newtxtext,newtxmath}
\usepackage[colorlinks=true,citecolor=red,linkcolor=black]{hyperref}
\makeatother

\title{Rainfall-runoff prediction using a Gustafson-Kessel clustering based Takagi-Sugeno Fuzzy model}

\author{Subhrasankha Dey*$^{1}$, Tanmoy Dam$^{2}$ \IEEEmembership{Student Member, IEEE}
\thanks{*Corresponding author}

\thanks{$^{1}$Subhrasankha Dey, Dept. of Infrastructure Engineering, The University of Melbourne, Parkville, Australia
        {\tt\small deys@student.unimelb.edu.au}}%
\thanks{$^{2}$ Tanmoy Dam, School of Engineering and Information Technology, University of New South Wales Canberra,  Australia
    {\tt\small t.dam@student.adfa.edu.au}}%
}

\begin{document}

\maketitle

\begin{abstract}
A rainfall-runoff model predicts surface runoff either using a physically-based approach or using a systems-based approach. Takagi-Sugeno (TS) Fuzzy models are systems-based approaches and a popular modeling choice for hydrologists in recent decades due to several advantages and improved accuracy in prediction over other existing models. In this paper, we propose a new rainfall-runoff model developed using Gustafson-Kessel (GK) clustering-based TS Fuzzy model. We present comparative performance measures of GK algorithms with two other clustering algorithms: (i) Fuzzy C-Means (FCM), and (ii) Subtractive Clustering (SC). Our proposed TS Fuzzy model predicts surface runoff using: (i) observed rainfall in a drainage basin and (ii) previously observed precipitation flow in the basin outlet. The proposed model is validated using the rainfall-runoff data collected from the sensors installed on the campus of the Indian Institute of Technology, Kharagpur. The optimal number of rules of the proposed model is obtained by different validation indices. A comparative study of four performance criteria: Root Mean Square Error (RMSE), Coefficient of Efficiency (CE), Volumetric Error (VE), and Correlation Coefficient of Determination (R) have been quantitatively demonstrated for each clustering algorithm. 
\end{abstract}


\section{Introduction}
A rainfall-runoff is an overland water flow due to precipitation happening over a drainage basin. A rainfall-runoff model predicts accumulated overland flow at an outlet of a drainage basin due to excess rainfall. A rainfall-runoff model can be developed using either a physically-based model or a data-driven black-box model. A physically-based model is constructed by the mass, momentum, and energy transformation equations such that the parameters of the model are directly related to the characteristics of a drainage basin (or a catchment area) \cite{kuiry2010coupled}. However, such models ignore frequent topographical changes of a catchment due to urbanization/human interventions and demand additional data (e.g. initial soil moisture, land use, evaporation and infiltration data, distribution, and rainfall duration) to improve model accuracy \cite{jacquin2006development}. In contrast to physically-based approaches, data-driven models can handle the topographic variation of a catchment and can produce output even with the absence of additional parameters and data, those are needed by a physically-based model \cite{morales2021self}. 

Recently, Fuzzy rule-based models are becoming a popular data-driven approach for rainfall-runoff modelling  \cite{jacquin2006development,lohani2005development,lohani2006takagi,hundecha2001development,morales2021self}. One of the important Fuzzy rule-based rainfall-runoff models is the Neuro-Fuzzy system, known as Adaptive Network-based Fuzzy Inference System (ANFIS) \cite{Jang93anfis}. ANFIS has been shown superiority when compared with other data-driven models such as Artificial Neural Network (ANN), Auto-Regressive Moving Average (ARMA), and Auto-Regressive with exogenous inputs (ARX) models. However, ANFIS is reported to be computationally expensive and tends to be over-fitting \cite{chang2017choice,morales2021self}. 

Performance of the existing Fuzzy rule-based models varies with the topographic nature of a catchment \cite{morales2021self}. Further, the models often suffer from uncertainty in the selection of lag-time, variability of the training data in the presence of many influencing parameters of a hydrological process. Hence, a comparative performance measure is adapted by researchers in order to achieve higher accuracy to predict runoff due to rainfall \cite{jacquin2006development,tabbussum2021performance}. 
Also, Takagi–Sugeno (TS) fuzzy systems remains a popular choice of capturing the non-linear relationships between rainfall and runoff in a catchment \cite{jacquin2006development,tabbussum2021performance}. 

The main contributions of this paper are as follows,
\begin{itemize}
  \item Gustafson-Kessel's (GK) clustering method has been adapted to learn the fuzzy structure, mainly premise and consequence parameters in TS fuzzy model identification. To obtain the optimal number of rules directly from the data, we used different cluster validating indices. 
  \item To validate the GK-based TS fuzzy model using the data obtained at a catchment area of Indian Institute of Technology, Kharagpur (an academic campus) \cite{dey2016real}. We evaluate our proposed model's performance in the multi-step-ahead runoff prediction scheme. We have also compared the performance with other existing clustering-based Fuzzy methods. 
\end{itemize}


\section{Literature review}
In earlier research, linear regression-based runoff prediction models are found  \cite{lohani2011comparative}. However, nonlinear dynamics of runoff due to rainfall were not considered and hence the model performance is not adequate in linear regression. In recent time, the following nonlinear data-driven approaches are becoming popular (i) Artificial Neural Network (ANN) \cite{dawson1998artificial,tayfur2006ann,lohani2011comparative} and (ii) Fuzzy rule-based Neuro-Fuzzy Model \cite{lohani2005development,lohani2006takagi,hundecha2001development,Chang2001Chen}.

The Fuzzy rule-based models are popularly based on the trial and error method to find out the optimal number of rules \cite{lohani2011comparative,nayak2005Fuzzy}. In a Fuzzy rule-based approach, membership functions and Fuzzy rules are determined with some degree of uncertainty which is present in a hydrological process \cite{morales2021self}. An improvement of a Fuzzy rule-based system is found in Neuro-Fuzzy models where a neural network is used to train the model's membership functions and Fuzzy rules \cite{nayak2004neuro,morales2021self}. Neuro-Fuzzy systems are often based on Takagi–Sugeno-Kang structure \cite{morales2021self}. ANFIS is a widely popular Neuro-Fuzzy system and is used in a number of data-driven rainfall-runoff modelling despite the overfitting issues of finding suitable membership functions \cite{nayak2004neuro,nayak2005b,mukerji2009flood,remesan2009runoff}. Other studies which have employed Neuro-Fuzzy Models (NFMs) for flow forecasting include DENFIS (Dynamic Evolving Neural-Fuzzy Inference System) models \cite{kasabov2002denfis}. Wavelet-based approaches are also integrated with the ANFIS model to handle the uncertainty in the observed data for removing the noise samples \cite{akrami2014development}. The effect of lag time on rainfall-runoff modeling by ANFIS has been studied earlier \cite{talei2012influence}. Additionally, an input selection method based on correlation and mutual information analysis is able to identify an optimum set of rainfall inputs for rainfall-runoff modeling by ANFIS \cite{talei2012influence}. Sometimes, non-sequential rainfall antecedents can produce better results compared to sequential rainfall inputs \cite{talei2012influence}. Hence, selection of the type and number of inputs are also important in a TS Fuzzy rule-based rainfall-runoff model \cite{govindraju2000artificial}. Rainfalls are considered by some researchers as the only inputs of the data-driven model \cite{chua2008comparison,talei2010aNovel}.  When the flow measurement is unavailable for sub-catchments in a catchment, rainfall data can be chosen as inputs to identify the more contributing sub-catchments \cite{chang2017choice}. However, the performance of ANFIS could be affected due to unnecessary complexity when so many inputs are involved \cite{talei2012influence}. The computational time can be significantly reduced using a fewer number of inputs. 

A combination of rainfall and runoff are also used as inputs to the Fuzzy models \cite{nayak2005Fuzzy,talei2013runoff} with increased computation complexity and overfitting errors. Some studies have suggested pruning of the unnecessary inputs by selecting a narrower time window around the most correlated rainfall antecedent with runoff \cite{nayak2007rainfall,nayak2005Fuzzy}. Hence, in this paper, we develop a simple four-input-based Takagi-Sugeno Fuzzy model to reduce computational complexity and reducing overfitting errors. We present a comparative study of three different Fuzzy clustering algorithms based the proposed TS Fuzzy model using real data collected at a pilot study area of a drainage basin. 


\section{General Structure of a TS Fuzzy model} \label{GKTSFM} 
In TS Fuzzy model, any nonlinear function is approximated by a set of linear models where each linear model is described as a rule \cite{takagi1985Fuzzy}. The structure of each rule is represented by premise and consequent variables. An input variable transformed into Fuzzy Membership Function (MF) is known as premise variable, whereas the
output of TS Fuzzy model is known as consequent variable. The consequent output is described by a linear functional relation between premise variables. In a TS Fuzzy model , each rule is described as follows,
\begin{align}
\nonumber {P^i}:\, IF \,&{x_{k1}} \,is \,{A_1}^i \,{\rm{AND}}\,IF \,{x_{k2}} \,is \,{A_2}^i ...\,{\rm{AND}}\,IF\,{x_{kn}}\,is\,{A_n}^i \\
\nonumber THEN\,{y_k}^i &= {a_0}^i + {x_k}_1.{a_1}^i + ... + {x_{kn}}.{a_n}^i\\
\nonumber &= {a_0}^i + \sum\limits_{j = 1}^n {{x_{kj}}.{a_j}^i},\,\, j = 1,2,..,n,\,i = 1,2,..,C\\
&= [1\,\,{x_k}^T]{\theta ^i}\, = \bar x{\theta ^i}
\label{TS_rule}
\end{align}
where $P^i$ is the structure of the rule $(i=1,2,...,C rule)$ and $n$ is the number of input variable. ${{\bf x}_k} = [{x_{k1}},{x_{k2}},...{x_{kn}}]$ be the input variables or premise variables of the model and $y_k$ is the consequent's output of $i^{th}$ rule. ${\theta ^i} = [{p_0}^i,{p_1}^i,...,{p_n}^i]\in {R^{(n+1)}}$ is the coefficients of consequent variable. The dumbbell shape MF (i.e. Gaussian MF) for each premise variable is considered here as,
\begin{align}
\nonumber {A_j}^i({x_{kj}}) = \exp [ - \frac{{{{({x_{kj}} - {\vartheta_j}^i)}^2}}}{{{{({\sigma _j}^i)}^2}}}] \in [{\rm{0}},{\rm{1}}], \\ 
\textrm{where, } (i = 1,2,..,C;j = 1,2,..,n) \label{gaussian_membership}
\end{align}
where $V =[{\vartheta _1}^i,{\vartheta _2}^i,...,{\vartheta _n}^i] \in {R^n}$ is the vector of mean of the GMF. The width is represented by $[{\sigma _1}^i,{\sigma _2}^i,...,{\sigma _n}^i] \in {R^n}$.

The final output of the TS Fuzzy model is represented by weighted average of each individual firing rule. Let, the system input variables and output variables are, ${x_k} = [{x_k}_1,{x_k}_2,...,{x_k}_n]{\,^T}(k = 1,2,...,N^{th}$ observation$)$ and ${y_k}(k = 1,...,N)$ respectively. Therefore, the output of TS Fuzzy model is given by,
\begin{equation}
{{{\hat y}_k} = \frac{{\sum\limits_{i = 1}^C {{w_k}^i.{y_k}^i}}}{{\sum\limits_{i = 1}^C {{w_k}^i} }}}
\label{Final TSoutput}
\end{equation}
where, the membership value (${A_j}^i({x_{kj}})$)of $j^{th}$ premise variable of $k^{th}$ sample is defined in (\ref{gaussian_membership}). The overall truth value (${{w_k}^i}$) of each rule can be calculated by minimum operator or logic AND operation. The truth value of $k^{th}$ sample is expressed by,
\begin{align}
\nonumber {w_k}^i = \mathop {\min }\limits_j \{ {A_j}^i({x_{kj}})\} , \\
{(i = 1,2,...,{\mkern 1mu} C{\mkern 1mu} ;{\mkern 1mu} {\mkern 1mu} j = 1,2,...,{\mkern 1mu} n\,;\,k = 1,2,...,N)} \label{truth value}
\end{align}

\section{Fuzzy clustering based TS Fuzzy model}
\label{Clustering_based_partition matrix} 
\subsection{Fuzzy partition matrix from the model data}
TS Fuzzy model can be designed from the input-output data by Fuzzy partition matrix. Let, the data vector is represented as, $D = \{{Z_k}:{Z_k} = {[{x_k}^T,{y_k}]^T},{x_k} = {[{x_{k1}},{x_{k2}},...,{x_{kn}}]^T},k = 1,2,...,N\}$. Hence, the Fuzzy partition matrix in a data set for premise variables is represented by clusters with  certain degree of membership values. Each membership value for each cluster is a part of the Fuzzy partition matrix.

\textbf{Definition} Let, $\Im  = \{ {Z_1},{Z_2},...,{Z_N}\} $ be a finite set and $2 \le C < N$ be number of clusters, then Fuzzy partition matrix for set $\Im $, is defined as

${U_{fc}} = \{ U \in {R^{C \times N}}|{\mu _{ik}} \in [0,1];\,\,\sum\limits_{i = 1}^C {{\mu _{ik}} = 1;} \,\,0 \prec \sum\limits_{k = 1}^N {{\mu _{ik}} < N,\,\,\forall } i\} $

where, ${\mu _{ik}}(i = 1,2,...,C\,and\,k = 1,2,...,N)$ be Fuzzy membership value of ${Z_k}$ sample is belonging to $\emph{i}^{th}$ cluster.

\subsection{GK clustering algorithm based Fuzzy partition matrix}
The whole input-output space is divided by Fuzzy clustering algorithm to achieve the Fuzzy partition matrix. Therefore, many clustering algorithms based TS Fuzzy models are found in \cite{chiu1994Fuzzy}, \cite{abonyi2002modified}, \cite{dam2016interval}. FCM clustering algorithm is popular of its kind \cite{dam2014ts}. The modified version of FCM algorithm is known as GK (Gustafson-Kessel) Algorithm \cite{babuka2002improved}. Cluster covariance values of each class is updated by the norm inducing distance metric. The geometrical shape provided by GK clustering algorithm is elliptical shape in nature whereas the FCM gives hyper-spherical shape. The objective function of GK algorithm is given by,
\begin{equation}
{\begin{array}{l}
{J_{GK}}(D;U,{\rm{\theta }},{\rm{v,}}\,{M_i}) = \sum\limits_{i = 1}^C {\sum\limits_{k = 1}^N {{\mu _{ik}}^m} } {G_{ik}}^2\\
subject\,to,\,\,\sum\limits_i^C {{\mu _{ik}}}  = 1
\end{array}}
\label{GK_objective}
\end{equation}
where the distance metric is ${G_{ik}}^2 = {[{Z_k} -{\textbf{v}^i}]^T}{M_i}[{Z_k} - {\textbf{v}^i}]$, ${{\bf{v}}^i}\in {R^{(n + 1)}}$ is cluster prototype in the input-output product space and ${M_i}$ is symmetric positive definite matrix, represents as volume size of $i^{th}$ cluster. Equation (\ref{GK_objective}) is a constrained optimization problem. Applying the Lagrangian multipliers ($\lambda$) to convert it into an unconstrained optimization form,
\begin{equation}
{\begin{array}{l}
L(U,\lambda ,{\rm{\theta }},{\rm{v}}) = {J_{GK}}(D;U,{\rm{\theta }},{\rm{v,}}\,{M_i}) - \sum\limits_{i = 1}^C {{\beta _i}} (|{M_i}| - \rho )\\
\,\,\,\,\,\,\,\,\,\,\,\,\,\,\,\,\,\,\,\,\,\,\,\,\,\,\,\,\, -
\sum\limits_{k = 1}^N {{\lambda _k}(\sum\limits_{i = 1}^C {{\mu
_{ik}} - 1} )}
\end{array}}
\label{lagrangian form}
\end{equation}

For minimization of (\ref{lagrangian form}) with the respect to ${\mu _{ik}}$, assuming ${G_{ik}}^2 \succ 0$, the final expression is as follows,
\begin{equation}
{{\mu _{ik}} = \frac{1}{{\sum\limits_{q = 1}^C {{{\left({\frac{{{G_{ik}}^2}}{{{G_{qk}}^2}}} \right)}^{\frac{1}{{m - 1}}}}}}}} \label{final miu}
\end{equation}
Similarly, the cluster center ($v^i$) is obtained by taking derivative of (\ref{GK_objective}) w.r.t. $ {\textbf{v}^i}$ as,
\begin{equation}
{\begin{array}{l}
\frac{{\partial L(...)}}{{\partial {{\rm{v}}^i}}} = \sum\limits_{k = 1}^N {{\mu _{ik}}^m[{x_k} - {{\rm{v}}^i}]}  = 0\\
 \Rightarrow \,\,\,{{\rm{v}}^i} = \frac{{\sum\limits_{k = 1}^N {{\mu _{ik}}^m{Z_k}} }}{{\sum\limits_{k = 1}^N {{\mu _{ik}}^m} }}
\end{array}}
\label{cluster center}
\end{equation}
Taking partial derivative (\ref{lagrangian form}) w.r.t $M^i$,
\begin{equation}
{\begin{array}{l}
\frac{{\partial L(...)}}{{\partial {M_i}}} = \sum\limits_{k = 1}^N {{\mu _i}{{_k}^m}[{{\rm{Z}}_k} - {{\rm{v}}^i}]} {[{{\rm{Z}}_k} - {{\rm{v}}^i}]^T}\\
\,\,\,\,\,\,\,\,\,\,\,\,\,\,\,\,\,\,\,\,\,\,\,\,\,\,\,\,\, - {\beta _i}\left| {{M_i}} \right|{M_i}^{ - 1} = 0\\
 \Rightarrow {M_i}^{ - 1} = \frac{{\sum\limits_{k = 1}^N {{\mu _i}{{_k}^m}[{{\rm{Z}}_k} - {{\rm{v}}^i}]} {{[{{\rm{Z}}_k} - {{\rm{v}}^i}]}^T}}}{{{\beta _i}\left| {{M_i}} \right|}}\\
\,\,\,\,\,\,\,\, = \sqrt[n]{{(\frac{1}{{\rho \left| {{F_i}}
\right|}})}}\,\,\,{F_i}.
\end{array}}
\label{Fuzzy covariance matrix}
\end{equation}
where $F_i$ is the covariance matrix. $\rho$ is a constant.

\begin{algorithm}
\caption{GK algorithm for a TS Fuzzy model}
 \begin{algorithmic}
 \State \textbf{Inputs}:
 \State Fuzziness parameter ($m =2$), Iteration number ( $l=0$).
   \State Termination constant ($\xi  = 0.001$)
   \State ${G_{ik}}^2$: Calculate the distance metric, ${G_{ik}}^2 = {[{Z_k} - {{\bf{v}}^i}]^T}({\rho }(det{({F_i})^{{\raise0.7ex\hbox{$1$} \!\mathord{\left/
 {\vphantom {1 n}}\right.\kern-\nulldelimiterspace}
\!\lower0.7ex\hbox{$n$}}}}){F_i}^{ - 1})[{Z_k} - {{\bf{v}}^i}]$ .
   \State ${U^0} \in {\mathbb{R}^{C \times N}}$: Fuzzy partition matrix for first \\iteration.
   \State \textbf{Output}:
     \State ${U^l}$: Fuzzy partition matrix for $l^{th}$ iteration.
     \State \textbf{Steps:}
     \State \textbf{1.} Calculate $v^i$ by using (\ref{cluster center}).
     \State \textbf{2.} Calculate covariance matrix $F_i  = \frac{{\sum\limits_{k = 1}^N {\mu _{ik} ^m \left( {Z_k  - v^i } \right)\left( {Z_k  - v^i } \right)^T } }}{{\sum\limits_{k = 1}^N {\mu _{ik} ^m } }}$.
     \State \textbf{3.} Update ${U^l}$ with distance metric ${G_{ik}}^2$.\\\
     \begin{equation}
     {{U^l} = \left\{ {\begin{array}{*{20}{c}}
{\frac{1}{{\sum\limits_{q = 1}^C {{{\left( {\frac{{{G_{ik}}^2({\theta ^i})}}{{{G_{qk}}^2({\theta ^q})}}} \right)}^{\frac{1}{{m - 1}}}}} }}}&{{\rm{If}}{\mkern 1mu} {G_{ik}}({\theta ^i}) > {\mkern 1mu} 0}\\
0&{{\rm{otherwise}}}
\end{array}} \right.}
\label{final miu Gk}
      \end{equation}
  \State until ${\left\| {{U^l} - {U^{l - 1}}} \right\|_2} \le \xi $ then stop; \\otherwise, $l=l+1$ and go to step 1.
    \end{algorithmic}
    \label{algo: GK}
\end{algorithm}

\subsection{Cluster validity index} \label{cluster validity index}
Cluster validity index provides a clear idea about the optimal number of partitions in the data space. The partition in the data space is obtained by clustering algorithms viz. FCM, GK and SC \cite{xie1991validity}. The optimal number of partitions (subspace) in a data space are obtained by varying the clustering numbers. However, varying the clustering numbers only may not be sufficient enough to provide the optimal numbers due to its dependency in the cluster shape. Therefore, a suitable validity index with proper partitioning algorithm is required to obtain the optimal numbers. Hence, six different validity measures are applied to find out the optimal number of clusters in the data space as follows: 

\subsubsection{Partition Coefficient (PC)}
The Partition Coefficient (PC) \cite{bezdek1974numerical}index measures of overlapping between clusters. It is defined by,
\begin{equation}
{{{V_{PC}} = \frac{{\sum\limits_{i = 1}^C {\sum\limits_{k = 1}^N{{{\left( {{\mu _{ik}}} \right)}^2}} } }}{N}}}
\label{PC}
\end{equation}
where ${\mu _{ik}}$ is the membership value of $k^{th}$ data point belonging to $i^{th}$ cluster. The PC method does not hold any connection between shape of data. It is concerned only with the partition matrix ($U$) i.e. it does not have any relation with data space. Therefore, this validity index may not favorable for highly complex dataset. The index provides an optimal number of clusters at a point that gives maximum value by varying the cluster number from $C=2$ to $C_{max}$.

\subsubsection{Partition Entropy (PE)}
The Partition Entropy (PE) index \cite{bezdek1974numerical} defines the fuzziness between partitions and is given by as, \begin{equation}
{V_{PE}} =  - \frac{{\sum\limits_{k = 1}^N {\sum\limits_{i = 1}^C{{\mu _{ik}}\log ({\mu _{ik}})}}}}{N}
\label{PE}
\end{equation}
where the minimization of ($V_{PE}$) provides the optimal number by varying cluster number from $C=2$ to $C_{max}$. Similarly, for the $V_{PE}$ index, that may not provide the exact number of partitions in a dataset because it does not hold any relation between the partitions shape.

\subsubsection{Modified Partition Coefficient (MPC)}
Both $V_{PC}$ and $V_{PE}$ indices monotonic decreasing tendency with varying $C$. The modified $PC$ (MPC) \cite{dave1996validating} can remove the monotonic decreasing tendency and is given by as,
\begin{equation}
{{{\rm{V}}_{{\rm{MPC}}}}{\rm{  =  1  -  }}\frac{C}{{C - 1}}{\rm{(1 - }}{{\rm{V}}_{{\rm{PC}}}}{\rm{)}}} \label{MPC}
\end{equation}
The optimal number of clusters is obtained by maximizing $V_{MPC}$ over the cluster number, varying from $C=2$ to $C_{max}$.

\subsubsection{Partition Index (SC)}
This index \cite{bensaid1996validity} is obtained by ratio of the compactness and separation of a cluster. The summation of the individual cluster validity measure is normalized by dividing cardinality of each cluster. The index is given by,
\begin{equation}
{{V_{SC}} = \sum\limits_{i = 1}^C {\frac{{\sum\limits_{k = 1}^N
{{{\left( {{\mu _{ik}}} \right)}^2}{{\left\| {{Z_k} - {v^i}}
\right\|}^2}} }}{{{N_i}\sum\limits_{j = 1}^C {{{\left\| {{v^j} -
{v^i}} \right\|}^2}} }}}} \label{SC}
\end{equation}
A lower value of the index indicates the optimal numbers.

\subsubsection{Separation Index (S)}
This index \cite{bensaid1996validity} is generally do the opposite effect of SC index. Minimum separation provides the optimal numbers of clusters. It is given by as,
\begin{equation}
{{V_S} = \frac{{\sum\limits_{i = 1}^C {\sum\limits_{k = 1}^N {{{\left( {{\mu _{ik}}} \right)}^2}{{\left\| {{Z_k} - {v^i}} \right\|}^2}} } }}{{{N_i}\mathop {\min }\limits_{i,j} {{\left\| {{v^j} - {v^i}} \right\|}^2}}}}
 \label{S}
 \end{equation}

\subsubsection{Xie-Beni (XB)}

The Xie-Beni index \cite{xie1991validity} is based on the compactness of the cluster and minimum separation variation between the clusters.
 The optimal value is obtained from the minimum values of the two ratios by varying the cluster number from $C=2$ to $C_{max}$.
 The mathematical expression for $V_{XB}$ is given by,
\begin{equation} {{V_{XB}} =
\frac{{\sum\limits_{i = 1}^C {\sum\limits_{k = 1}^N {{{\left(
{{\mu _{ik}}} \right)}^2}{{\left\| {{Z_k} - {v^i}} \right\|}^2}} }
}}{{N\mathop {\min }\limits_{i,j} {{\left\| {{v^j} - {v^i}}
\right\|}^2}}}}
 \label{XB}
 \end{equation}

\subsection{Parameter estimation}
In this section, optimal number of rule-base Fuzzy partition matrix obtained by GK algorithm is used to estimate the premise parameters. Here, the Gaussian Membership Function (GMF) for premise variables is considered. The mean and the width (standard deviation) \cite{dam2017clustering, dam2016interval, dam2014ts, dam2015block, dam2015interval} are
obtained as follows,

\begin{equation}
{{\vartheta _{\rm{j}}}^i = \frac{{\sum\limits_{k = 1}^N {{\mu
_{ik}}^m{{\rm{x}}_{kj}}} }}{{\sum\limits_{k = 1}^N {{\mu _{ik}}^m}
}}} \label{Mean of GMF}
\end{equation}
\begin{equation}
{{\sigma _j}^i = \sqrt {\frac{{2*[\sum\limits_{k = 1}^N {{\mu
_{ik}}^m{{({x_{kj}} - {\vartheta _j}^i)}^2}]} }}{{\sum\limits_{k =
1}^N {{\mu _{ik}}^m} }}} \,\,} \label{variance of GMF}
\end{equation}

The premise model parameters are identified from the Fuzzy partition matrix $(U)$ and model data $(Z)$. The coefficients of consequent part are identified. The obtained premise parameters are applied to make a global matrix form of consequent coefficients. The matrix form of coefficients is expressed as,
 \begin{equation}
{Y = \pi \zeta  + \varepsilon }
 \label{Global matrix form}
 \end{equation}
where $\pi$ is the regression vector that is derived from the input vector and truth value $(w)$. $Y = [{y_1},{y_2},...,{y_N}]$ is the
observed data vector and $\varepsilon$ is the modeling error for $N^{th}$ number of observations. The global form of the consequent coefficients is $\zeta=[{p_0}^1,{p_1}^1,...,{p_n}^1,{p_0}^2,{p_1}^2,...,{p_n}^2,...,{p_0}^C,{p_1}^C,...,{p_n}^C]$. The regression vector is represented as,
\begin{equation}
{\pi ({x_k}) = [{\nabla _{1k}}\,\,{\nabla_{1k}}{x_{k1}}\,\,{\nabla _{1k}}{x_{k2}}...\,{\nabla_{1k}}{x_{kn}}\,\,\,...{\nabla_{Ck}}\,...\,{\nabla_{_{Ck}}}{x_{kn}}]}
\label{regression vector}
\end{equation}
where, the truth value is obtained as follows,
\begin{equation}
{{\nabla _{ik}}\, = \frac{{{w_k}^i}}{{\sum\limits_{i = 1}^C{{w_k}^i} }}} \label{truth value_weight}
\end{equation}

The matrix inversion of Equation (\ref{Global matrix form}) may not feasible for all the time to determine the coefficients of $\zeta$ \cite{dam2016interval}. Therefore, the Orthogonal Least Square (OLS) has been applied to the Equation (\ref{Global matrix
form}) to identify the coefficients of consequent parameters \cite{dam2014ts}.

\section{Proposed TS Fuzzy model}
\label{data structure}
The selection of input and output variables is the first step to develop a TS-Fuzzy rainfall-runoff model. Runoff at the outlet of a drainage basin is directly related to previous values of runoff and rainfall at different locations. However, the accumulated water due to rainfall takes a variable time to reach the exit of the basin depending on the catchment topography. The time interval from the center of mass of rainfall excess to the peak of the resulting hydrograph is known as lag-time. The effect of lag-time has been removed from the output data-set by adding the same delay in all the input variables. We define our proposed model after (removing the lag-time in the training dataset) as $M_4$ model (with four input variable), having the following data structure:

\begin{equation}
{{M_4}:\,\,\,\,{y_k} = f({y_{k -1}},x_k^1,x_k^2,x_k^3)}
 \label{TSFM_M4}
\end{equation}
where $y_k$ is the observed runoff and $x_k^1$, $x_k^2$ and $x_k^3$ are the rainfall collected at different locations at the $k^{th}$ time interval of a time horizon ${0,T}$ hours. Let $\delta^{0}_{pred}$ = time difference between $k^{th}$ and ${k-1}^{th}$ data point. We define the set $\delta_{pred} = {\delta^{0}_{pred},...,\delta^{i_N}_{pred},...}$ such that $\delta^{i_N-1}_{pred} = i_N \cdot \delta^{0}_{pred}$ and so on, where $i_N$ is an integer. A subset of $\delta_{pred}$ is called as a time ahead runoff prediction scheme.   

\subsection{Flow chart of a GK algorithm based TS Fuzzy model}

\begin{figure}[h!] \centering
\includegraphics[width=0.5\textwidth]{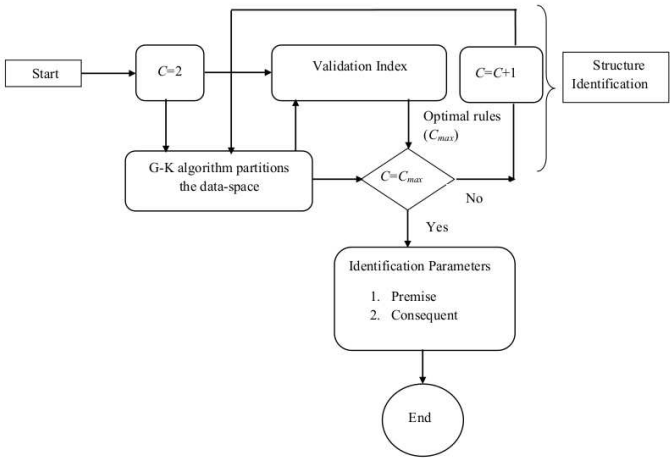}
\caption{Architecture of the GK algorithm based TS Fuzzy model}
\label{flowchart}
\end{figure}

Figure \ref{flowchart} describes the general architecture of the proposed GK algorithm based TS Fuzzy model. The model architecture is broadly classified into the following: 
\paragraph{{Finding optimal rule-base:} The GK clustering algorithm (Algorithm\ref{algo: GK}) along with validation index has been applied to input-output data set to create optimal Fuzzy subspaces (Fuzzy rules). Each optimal Fuzzy subspace is associated to a Fuzzy rule i.e. the output of optimal Fuzzy  ubspaces is described by Fuzzy partition matrix (${U}$). Once the optimal subspace is obtained from the both algorithms, then obtained $U$ has been used to estimate the model parameters.}

\paragraph{{Estimating TS Fuzzy model parameters:} Gaussian membership function parameters (mean and width) are obtained by using eq.(\ref{Mean of GMF}) and eq.(\ref{variance of GMF}) respectively. We have formulated a global matrix form of regression vector to identify the consequent parameters. OLS is applied to eq.(\ref{Global matrix form}) estimate the consequent parameters.}

\section{Data and validation technique}\label{DATAcollection}
\begin{figure}[h] \centering
{\includegraphics[width=0.5\textwidth]{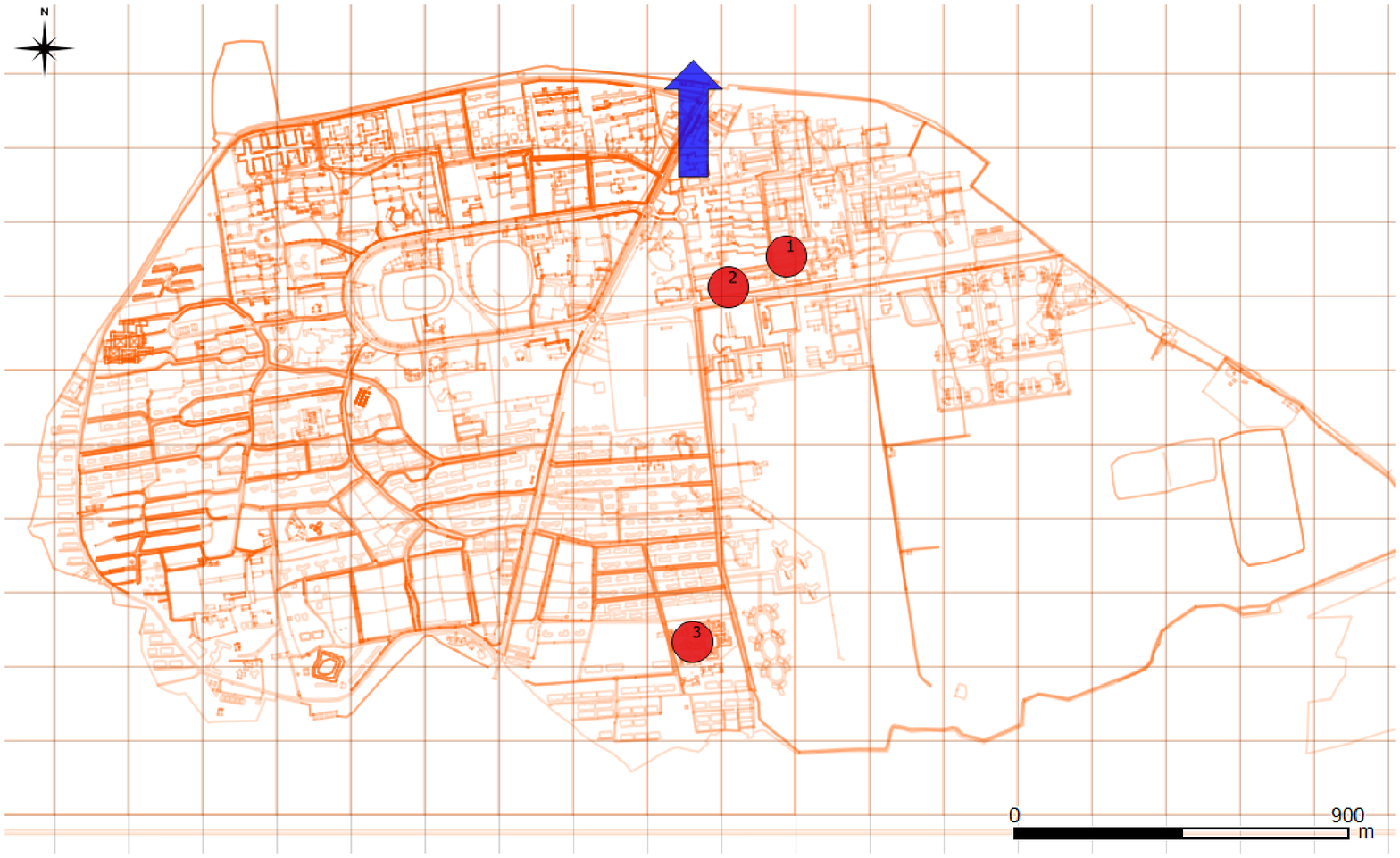}}
\vspace{+1em}
$\CIRCLE$ - Rainfall measuring nodes;
$\Uparrow$ - Water-level measuring node
\caption{Actual locations of spatially distributed sensing nodes deployed in the pilot study area} \label{IITKGP_location}
\end{figure}

A part of the campus of IIT Kharagpur is used as a pilot test-bed for the data acquisition process as shown in Figure \ref{IITKGP_location}. The validation data are obtained from an real-time web-based hydro-meteorological sensor network platform that exists in the study area \cite{dey2016real}. In Figure \ref{IITKGP_location}, the red markers are indicating locations of tipping bucket rain-gauges for on-line measurement of rainfall, and the blue up arrow indicates outlet of the drainage basin, where a rectangular weir with pressure type digital water depth sensor is installed to calculate pressure head (or water level). Pressure head is converted to surface runoff flow using Equation 1, 2 and 3 of  \cite{dey2016real}. 

A rainfall-runoff data set collected from the storm event in the $05^{th}$ May, $2015$ between $21:12$ Hours to $23:52$ Hours, comprising of $302$ observations (Data Set 1) is used to train the model parameters. Collected pressure head data from the storm event in the $22^{nd}$ September, $2015$ between $21:40$ Hours to $23:40$ Hours, comprising of $273$ observations (Data Set 2) are used to validate the models. Each observation used for training and validation purpose is comprises of rainfall data obtained from three spatially distributed location and pressure head data obtained at the exit of the study area. In Data set 1 (training data) and Data set 2 (validation data), collected data are taken at $30$ seconds interval. Hence, the minimum time difference between each observation $\delta^{0}_{pred}$ is $30$ seconds in Data Set 1 and Data Set 2. 
The proposed model is trained with Data Set 1 and validated using Data Set 2 without changing the dimension of each data point. Next, we normalise the input output dataset to develop a dimensionless training and validation data set. The models are labelled with a parenthesis $(N)$ when dimensionless input and output data set used to calculate the performance measure. The model performance is validated through statistical and hydrological performance measures such as Root Mean Square Error (RMSE), Coefficient of Efficiency (CE), Volumetric Error (VE) and correlation coefficient of determination (R) between observed data and simulated data. The model performance criteria are given by,
\begin{equation}
{\begin{array}{l}
RMSE = \sqrt[2]{{\frac{{\sum\limits_{k = 1}^N {{{({y_k} - {{\hat y}_k})}^2}} }}{N}}}\\
{CE} = \left( {1 - \frac{F}{{{F_0}}}} \right);\\
VE = \left( {\frac{{\sum {{y_k} - \sum {{{\hat y}_k}} } }}{{\sum {{y_k}} }}} \right) \times 100\\
R = \frac{{\sum\limits_{k = 1}^N {({y_k} - \bar y)({{\hat y}_k} -
{{\bar y}_p})} }}{{\sqrt[2]{{\sum\limits_{k = 1}^N {{{({y_k} -
\bar y)}^2}} }}\sqrt[2]{{\sum\limits_{k = 1}^N {{{({{\hat y}_k} -
{{\bar y}_p})}^2}} }}}};{{\bar y}_p} = \frac{1}{N}\sum\limits_{k =
1}^N {{{\hat y}_k}}
\end{array}}
\label{performance_expression}
\end{equation}
where ${y_k}$ is the observed pressure head (in mm), ${{\hat y}_k}$ is the predicted pressure head from the model. ${\bar y}$ and ${{\bar y}_p}$ are the mean of observed and predicted data respectively. The RMSE performance measure shows the algorithm capability for predicting the observed data. Lower values of RMSE are indicating better fit with the observed data. The $CE$ performance measure indicates the quality of the model verification data taken for different lengths over different time interval. The correlation coefficient ($R$) is used to check the goodness-of-fit of the model. 

\section{Results and Discussion} \label{Results and discussion}
We discuss two types of errors: (i) prediction error i.e. error observed in the next immediate observation, and (ii) time ahead runoff prediction error i.e. error observed in a particular time ahead runoff prediction scheme.   
 
\subsection{Prediction error comparison} \label{DATA SET 2}

\begin{table}  
\centering\caption{Prediction error comparison for all the different algorithm based $M_4$ model in both dimensional and dimensionless input-output setting} \label{GK based Fuzzy Model_DS2}
\begin{tabularx}{\columnwidth}{l|lXXX|lXXX|}
 \multicolumn{1}{l}{} & \multicolumn{4}{c}{Training} &
\multicolumn{4}{c}{Validation} \\ \hline
 Algorithm  & RMSE & VE & CE & R & RMSE       & VE & CE & R\\ \hline
 $GK$   &0.00 &0.00   &1.0   &1.00   &0.00      &0.00   &1.0    &1.0   \\ 
 $GK (N)$   &0.00  &0.002 &0.99   &0.99   &0.00  &0.002 &0.99   &1  \\  \hline
 $FCM$     &0.04 & 0.013  & 1   &1   &0.49 &0.05 & 1  & 1  \\ 
 $FCM (N)$   &0.02   &1.09   &0.99   &1  &0.02   &2.29   &0.99   &1  \\  \hline
 $SC$       & 0.03     & 0.01  &  1  & 1  &  0.06 &   0.01  & 1   & 1   \\ 
 $SC (N)$   &0.01   &1.12   &1  &1  &0.04   &2.96   &0.98   &0.99   \\  \hline
\end{tabularx}
\end{table}

GK, FCM and SC algorithms based $M_4$ TS Fuzzy models are constructed for both training (Data Set 1) and validation (Data Set 2). Table \ref{GK based Fuzzy Model_DS2}, specifies the performances of $M_4$. The performance of the $M_4$ model is listed in the Table \ref{GK based Fuzzy Model_DS2} for both dimensional and dimensionless input of the Data Set 2. The performances of six different cluster validation indices as presented in Section \ref{cluster validity index} have given minimum values at $3$ clusters. Hence, there are $3$ optimal number of clusters (rules) present in the model data set of $M_4$. Error measures for the six $M4$ models are presented in Table \ref{GK based Fuzzy Model_DS2} for both the training and validation data. In Table \ref{GK based Fuzzy Model_DS2}, the row that contains dimensionless input are identifiable by the '(N)' beside the algorithm (e.g. $GK (N)$). 

The FCM algorithm based $M_4$ model is generated by MATLAB "genfis3" function. The optimal number of rules in the FCM based Fuzzy structure are similar to the optimal number of rules, obtained by GK validation indices. Table \ref{GK based Fuzzy Model_DS2} indicates that FCM algorithm based model is not as good as GK algorithm while estimating the runoff data. The SC algorithm based $M_4$ model is generated by MATLAB "genfis2" function where number of rules in the model structure is similar to the optimal number of rules (provided by GK validation indices). Table \ref{GK based Fuzzy Model_DS2} indicates that SC based TS Fuzzy model has a better performance than that of FCM algorithm but it is not as good as the GK algorithm based $M_4$ model.


\begin{figure}[h!] \centering
\includegraphics[width=0.5\textwidth]{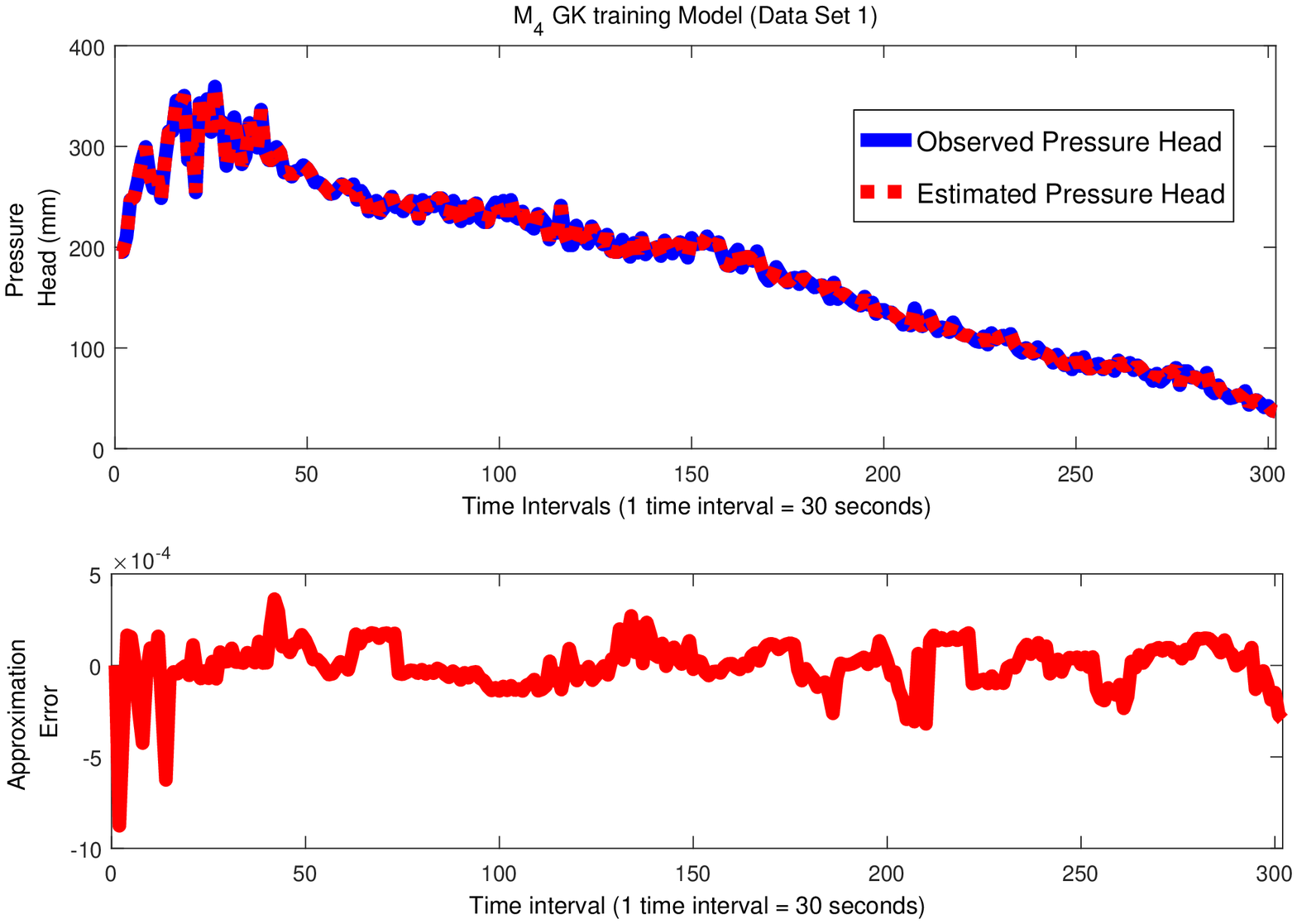}
\caption{Observed vs estimated pressure head for GK based $M_4$ model}
\label{GK4_train}
\end{figure}
\begin{figure}[h!] \centering
\includegraphics[width=0.5\textwidth]{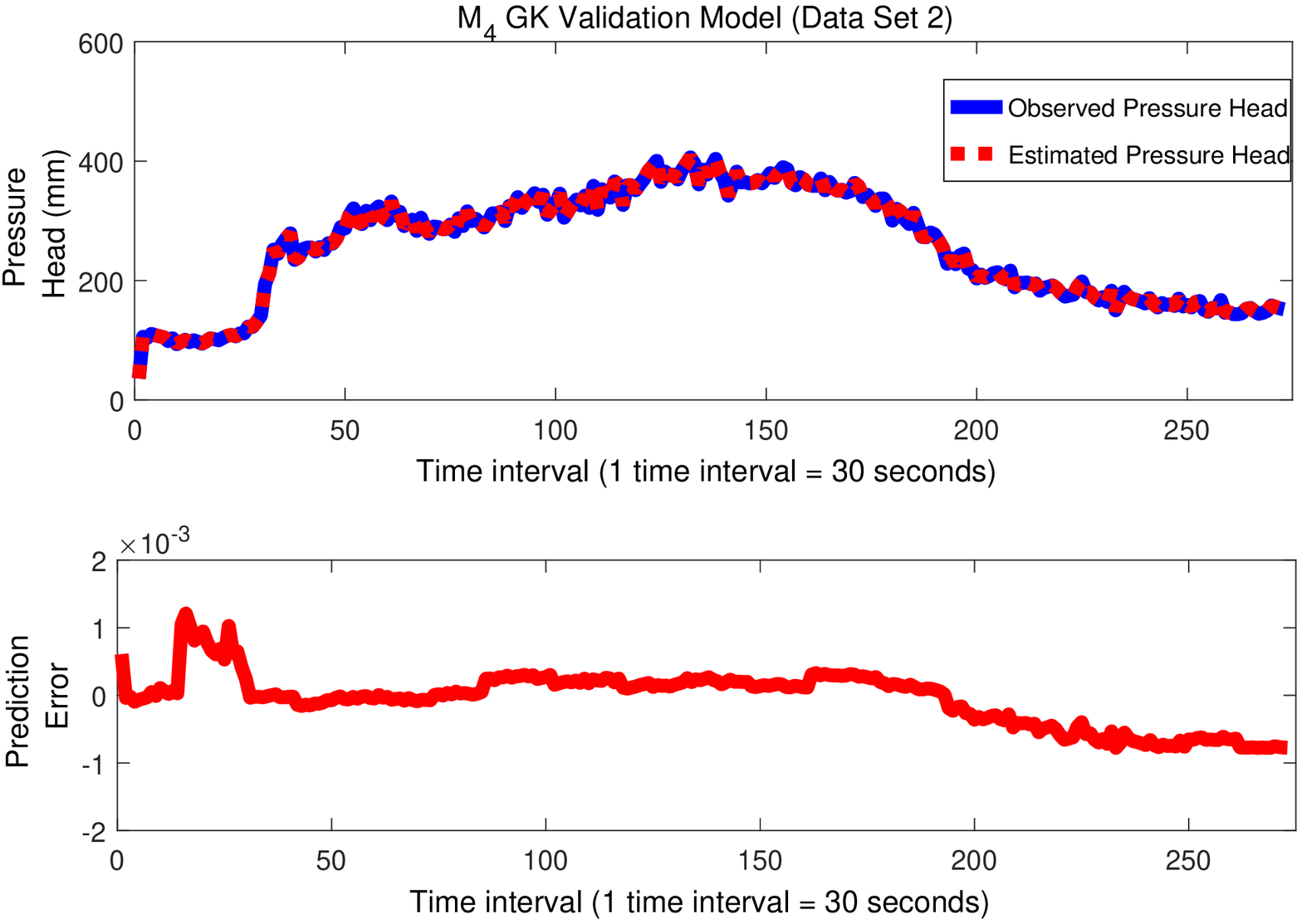}
\caption{Observed vs estimated pressure head for GK based $M_4$ model}
\label{GK4_test}
\end{figure}

\begin{figure}[h!] \centering
\includegraphics[width=0.5\textwidth]{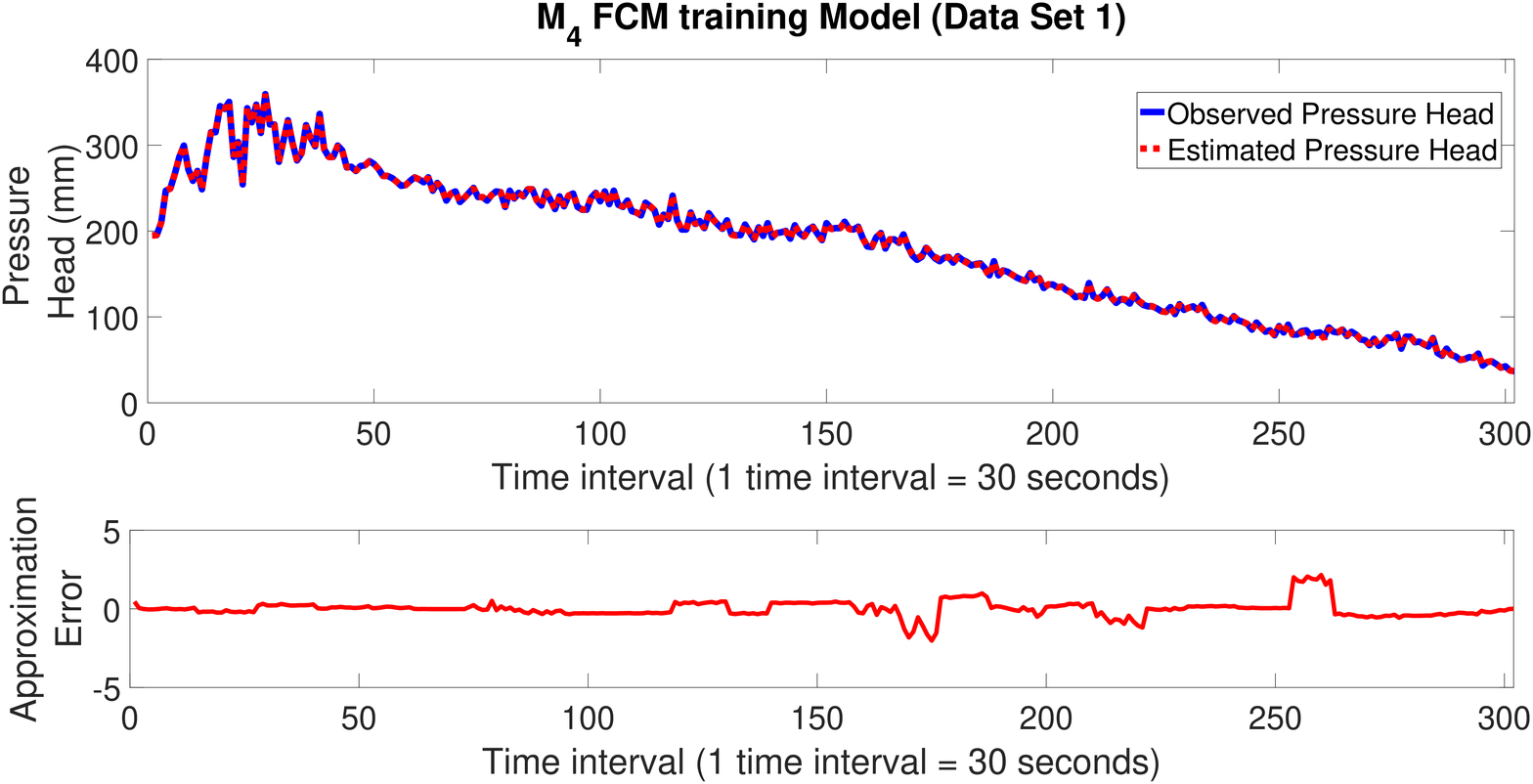}
\caption{Observed vs estimated pressure head for FCM based $M_4$ model}
\label{FCM4_train}
\end{figure}
\begin{figure}[h!] \centering
\includegraphics[width=0.5\textwidth]{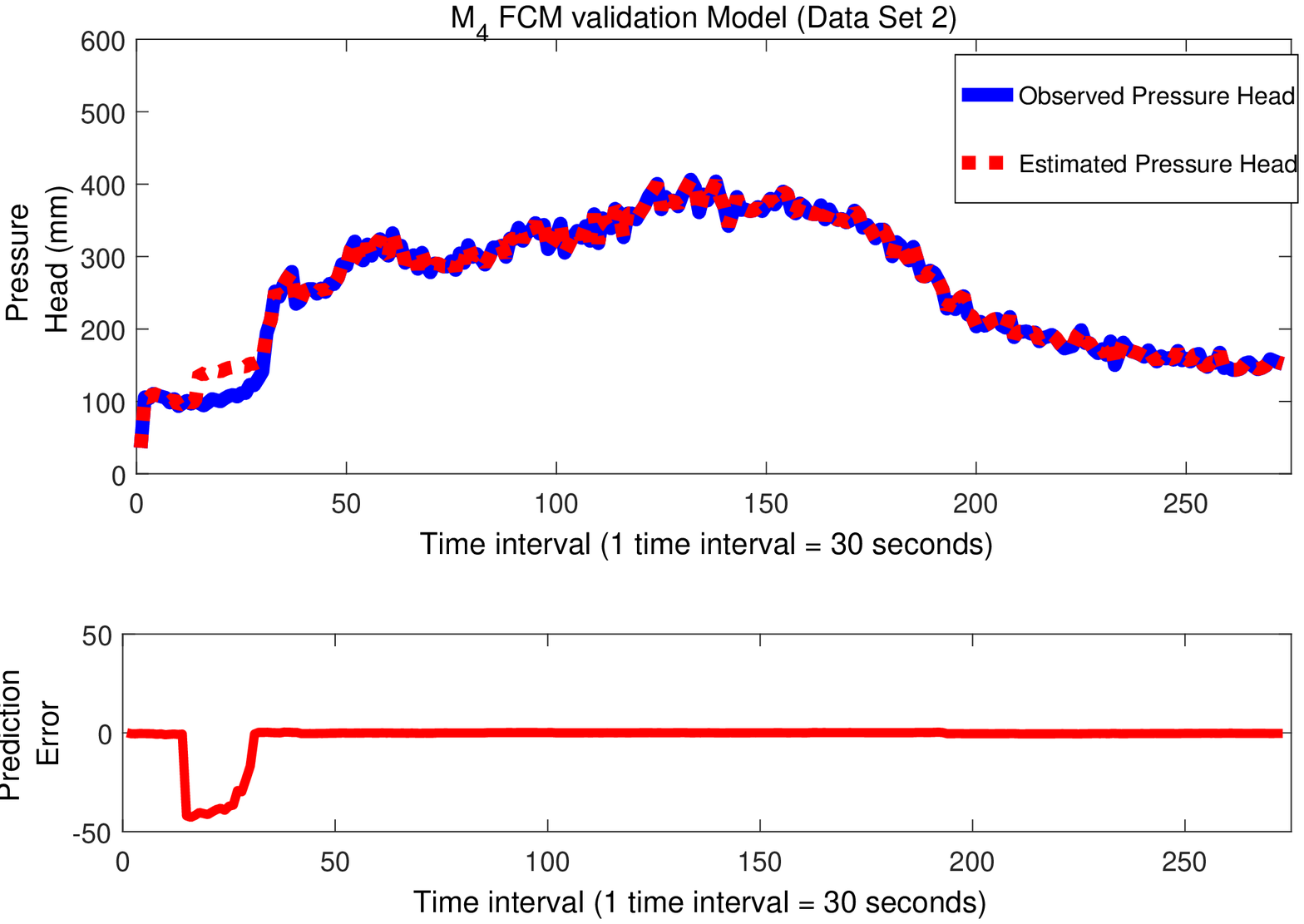}
\caption{Observed vs estimated pressure head for FCM based $M_4$ model}
\label{FCM4_test}
\end{figure}

\begin{figure}[h!] \centering
\includegraphics[width=0.5\textwidth]{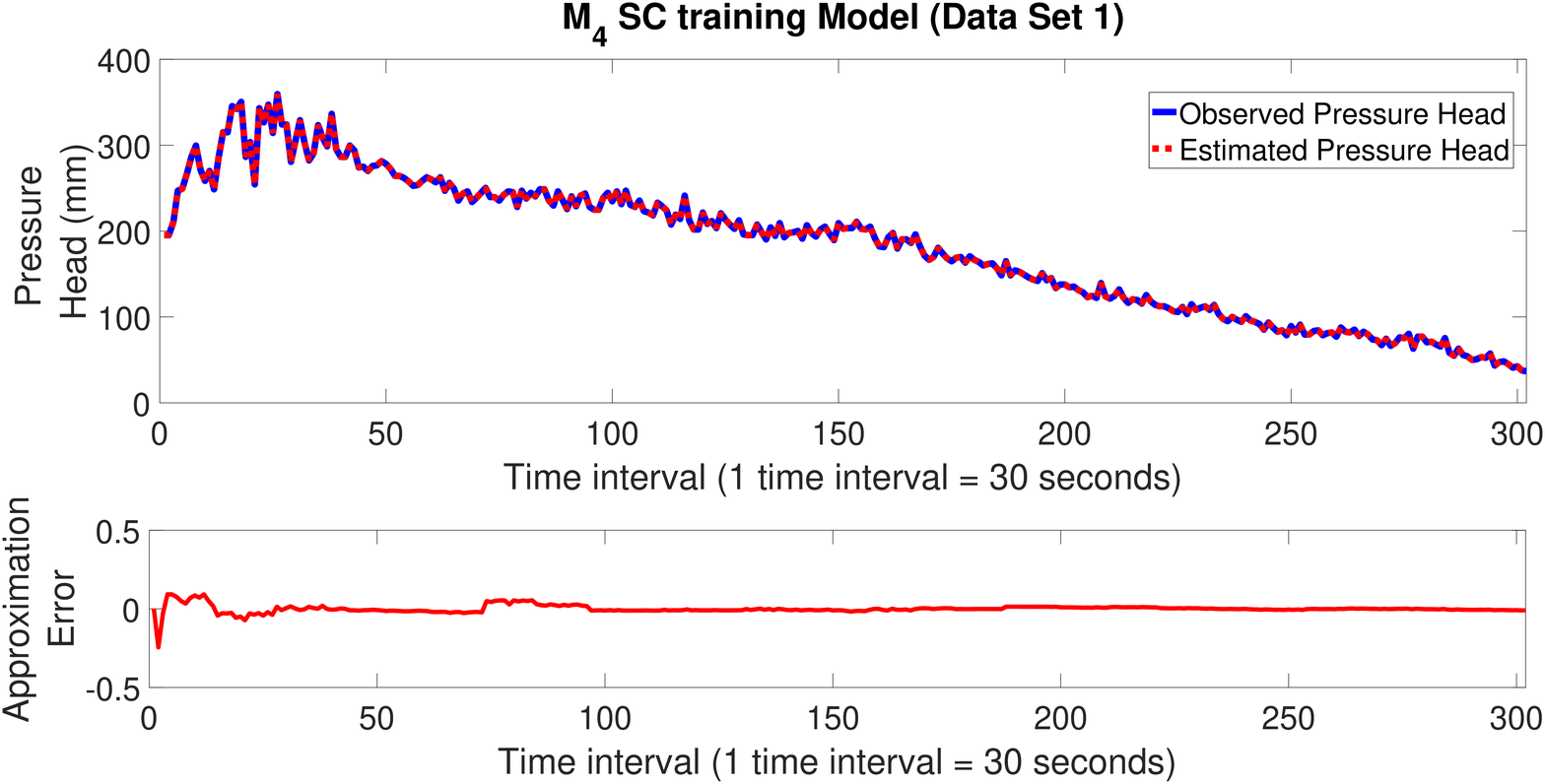}
\caption{Observed vs estimated pressure head for SC based $M_4$ model}
\label{subclus4_training}
\end{figure}
\begin{figure}[h!] \centering
\includegraphics[width=0.5\textwidth]{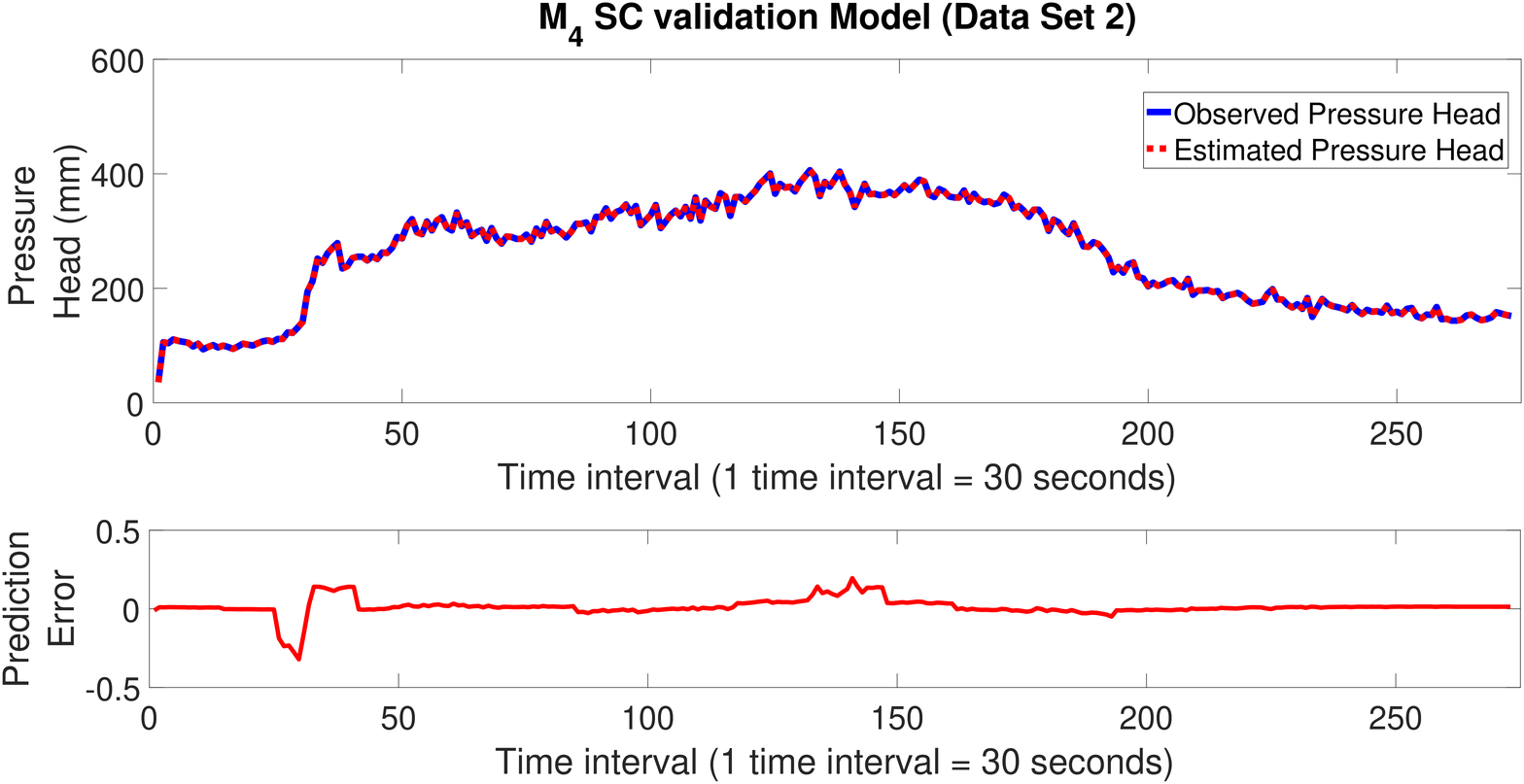}
\caption{Observed vs estimated pressure head for SC based $M_4$ model}
\label{sublcus4_testing}
\end{figure}

The GK based $M_4$ model performances for the training dataset and validation dataset are shown in the Figure \ref{GK4_train} and Figure \ref{GK4_test}. The training and testing dataset performances by SC based $M_4$ model are shown in the Figure \ref{subclus4_training} and Figure \ref{sublcus4_testing} respectively. The FCM based $M_4$ model performances for the training and testing dataset are shown in the Figure \ref{FCM4_train} and Figure \ref{FCM4_test} respectively. 

\subsection{Time ahead runoff prediction error comparison}

This section presents an error performance comparison of the algorithmic performances of the $M_4$ model when the data is predicted according to three different prediction ahead time i.e. three different schemes selected from the set $\delta_{pred}$. We demonstrate a comparison between  $\delta^1_{pred} = 60$ seconds, $\delta^4_{pred} = 150$ seconds and $\delta^9_{pred} = 300$ seconds in Table \ref{Error_comparison}. We should note that Table \ref{GK based Fuzzy Model_DS2} represents the errors obtained in the time ahead prediction scheme $\delta^0_{pred} = 30$ seconds. 

\begin{table}[h!]\centering
\centering \caption{Error comparison for three different prediction ahead time} \label{Error_comparison}
\begin{tabularx}{\columnwidth}{l|lXXX|lXXX|}
 \multicolumn{1}{l|}{$\delta^1_{pred}$} & \multicolumn{4}{c|}{Training} &
\multicolumn{4}{c|}{Validation} \\
  & RMSE & VE & CE & R & RMSE       & VE & CE & R\\ \hline
 $GK$    &0.00 &0.00   &1  &1  &8.99 &0.3 &0.99&0.99
\\ 
 $FCM$    &0.93  &0.35 &0.99 &0.99&9.42&1.07 &0.99 &0.99  \\ 
 $SC$     &0.33  &0.05 &1  &1  &9.25   &0.31   &0.99 &0.99
  \\ \hline \hline
 \multicolumn{1}{l|}{$\delta^4_{pred}$} & \multicolumn{4}{c|}{Training} &
\multicolumn{4}{c|}{Validation} \\ 
  & RMSE & VE & CE & R & RMSE       & VE & CE & R\\ \hline
 $GK$    &0.00   &0.00    &1  &1  &12 &0.61 &0.99   &0.99 \\ 
 $FCM$    &0.26   &0.11 &1  &1  &13.01    &0.88 &0.98   &0.99 \\ 
 $SC$     & 0.07  & 0.02    & 1 & 1 & 12.94   & 0.63    & 0.98    & 0.99 \\ \hline \hline
 \multicolumn{1}{l|}{$\delta^9_{pred}$} & \multicolumn{4}{c|}{Training} &
\multicolumn{4}{c|}{Validation} \\ 
 & RMSE & VE & CE & R & RMSE       & VE & CE & R\\ \hline
 $GK$    &0.00   &0.00    &1  &1  &16.05    &1.12 &0.97   &0.99 \\ 
 $FCM$    &0.014   &0.004 &1  &1  &18.06 &1.23  &0.96  &0.99 \\ 
 $SC$     &0.006   &0.002 &1  &1  &18.05 &1.21 &0.96 &0.99\\ \hline
\end{tabularx}
\end{table}

\begin{figure}[h!] \centering
\includegraphics[width=0.5\textwidth]{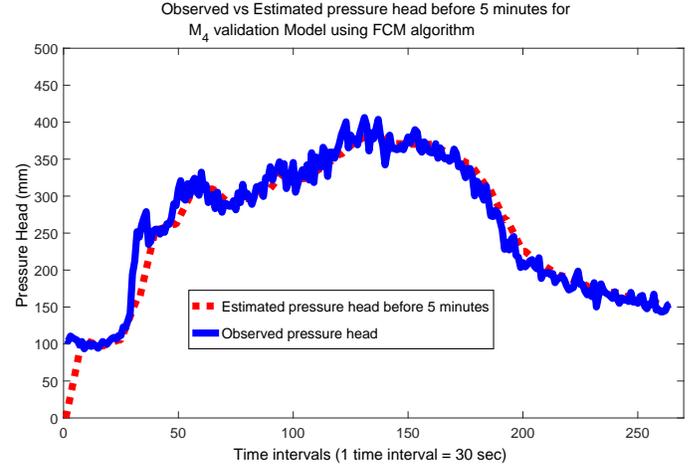}
\caption{Estimated pressure head before $5$ minutes of observed runoff for FCM based $M_4$ model}
\label{5com}
\end{figure}

Table \ref{Error_comparison} shows that the RMSE error of the $M_4$ model for GK algorithm has a $\pm16\%$ tolerance while predicting the runoff 10 times (300s/30s) earlier. The RMSE error of the $M_4$ model for FCM and SC algorithms have a $\pm18\%$ tolerance for the same prediction time interval. Figure \ref{5com} is showing the estimated pressure head before $5$ minutes of observed runoff for FCM based $M_4$ model that has the highest RMSE error among rest of the models. 

\section{Conclusion and Future Work}\label{conclusion}
This paper presents a GK clustering algorithm based TS fuzzy model to predict runoff due to rainfall using read world observations. Further, a comparative study between FCM and SC clustering algorithms is developed with various error measures. The proposed model is trained and validated with a different set of observations recorded on different days. Validation results show that the GK algorithm based $M_4$ model performs significantly better than FCM and SC algorithms. GK algorithm also performs better in the time ahead runoff prediction error that increases with $i_N$ (the superscript of $\delta^{i_N}_{pred}$).

Proposed models have been trained and validated over a limited number of rainfall-runoff data due to the catchment geography and climate characteristics. As future work, the methodologies described in this paper may be applied to those catchments where rainfall-runoff data is available for a longer duration of time to predict the surface runoff due to rainfall. Other hydrological inputs (e.g. Antecedent moisture conditions) and/or an increasing number of spatially distributed rainfall stations may be integrated with the input data set for a better prediction of the output.


\bibliographystyle{IEEEtran}

\bibliography{mybibfile}

\end{document}